\documentclass[journal]{IEEEtran}

\usepackage{url}

\usepackage{times}
\usepackage{epsfig}
\usepackage{graphicx}
\usepackage{amsmath}
\usepackage{amssymb}
\usepackage{algorithm}
\usepackage{algorithmicx}
\usepackage{algpseudocode}
\usepackage{bm}
\usepackage{caption}

\usepackage{url}
\usepackage{booktabs}
\usepackage{multirow}
\usepackage{multirow}
\usepackage{wrapfig}
\usepackage{wrapfig,lipsum,booktabs}
\usepackage{graphicx}
\usepackage{array}
\usepackage{color, colortbl}
\usepackage{array,hhline}
\usepackage[]{hyperref}

\newif\ifshowfig\showfigfalse

\ifCLASSINFOpdf
\else
\fi
\begin{document}
%

\title{SSFG: Stochastically Scaling Features and Gradients for  Regularizing  Graph  Convolutional Networks}

\author{Haimin~Zhang,
        ~Min~Xu,~\IEEEmembership{Member, IEEE},
        ~Guoqiang~Zhang,~\IEEEmembership{Member, IEEE}, ~Kenta~Niwa,~\IEEEmembership{Member, IEEE}

 \thanks{Haimin Zhang, Min Xu (\emph{corresponding author}) and Guoqiang Zhang are with the School of Electrical and Data Engineering, Faculty of Engineering and Information Technology, University of Technology Sydney, 15 Broadway, Ultimo, NSW 2007, Australia (Emails: Haimin.Zhang@uts.edu.au, Min.Xu@uts.edu.au, Guoqiang.Zhang@uts.edu.au).}
 \thanks{Kenta Niwa is with  the NTT Media Intelligence Laboratories, NTT Corporation, Chiyoda-ku 100-8116, Japan (Email: kenta.niwa@gmail.com)}
}
\maketitle

\begin{abstract}
  Graph convolutional networks have been successfully applied in various graph-based tasks.
  In a typical graph convolutional layer, node features are updated by aggregating neighborhood information.
  Repeatedly applying graph convolutions can cause the oversmoothing issue, \emph{i.e.},  node features at deep layers converge to similar values.
  Previous studies have suggested that oversmoothing is one of the major issues that restrict the performance of graph convolutional networks.
  In this paper, we propose a stochastic regularization method  to tackle the oversmoothing problem.
  In the proposed method, we stochastically scale features and gradients (SSFG) by a factor sampled from a probability distribution in the training procedure.
  By explicitly applying a scaling factor to break feature convergence, the oversmoothing issue is alleviated.
  We show that applying stochastic scaling at the gradient level is complementary to that applied at the feature level to improve the overall performance.
  Our method does not increase the number of trainable parameters.
  When used together with ReLU, our SSFG can be seen as a stochastic ReLU activation function.
  We experimentally validate our SSFG regularization method on three commonly used types of graph networks.
  Extensive experimental results on seven benchmark datasets for four graph-based tasks demonstrate that our SSFG regularization is effective in improving the overall performance of the baseline graph networks.
  The code is available at  \href{https://github.com/vailatuts/SSFG-regularization}{\color{blue}https://github.com/vailatuts/SSFG-regularization}.

\end{abstract}


\IEEEpeerreviewmaketitle

\section{Introduction} \label{sec:intro}

Data are organized in graph structures in various domains.
Social networks, citation networks, molecular  structures, protein-protein interactions---all of these domains can be modeled using graphs.
Developing powerful graph learning algorithms is important for many real-world applications such as recommendation systems \cite{wu2019session}, link prediction \cite{zhang2018link}, knowledge graphs \cite{chami2020low} and drug discovery \cite{klambauer2017self}.
Motivated by the success of deep convolutional networks, recent years have seen considerable interests in generalizing deep learning techniques to the  graph domain.

Compared with images and sequence data, graphs have a much complex topographical structure.
The nodes in a graph can have a very different number of neighbours, and there is no fixed node ordering  for a graph.
Early methods for learning on graphs are primarily based on recurrent neural networks.
These methods involve a process that iteratively propagates node features until the node features reach a stable point.
In recent years, graph convolutional networks (GCNs) that leverage graph convolutions have become the dominant approach for graph  learning.
Graph convolutions update node features by aggregating neighbourhood information.
Compared with recurrent-based methods, GCNs are much efficient in learning on graph-structured data.

\begin{figure}[!t]

  \begin{center}
  \includegraphics[width=.48\textwidth]{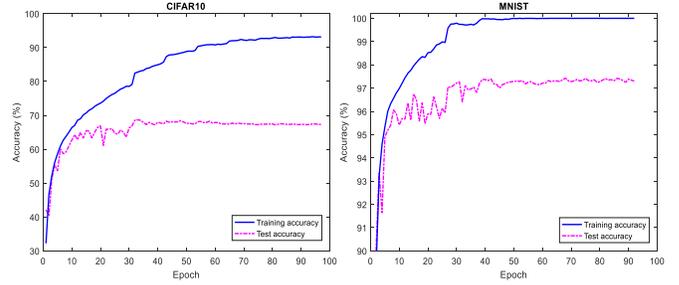} 
  \end{center}
  \caption{An illustration that shows the overfitting issue with graph networks for superpixel graph classification on CIFAR10 and MNIST \cite{dwivedi2020benchmarking}, the results are obtained using a four-layer GatedGCN. }
  \label{fig:illustration}
\end{figure}

While recently significant progress has been made in graph networks, overfitting and underfitting are major issues that restrict the performance of graph  networks.
Overfitting is more commonly seen in graph learning tasks, resulting in the model not generalizing well on unseen samples.
This issue usually comes with over-parametrization.
An illustration of this phenomenon is show in Fig.  \ref{fig:illustration}, we see that the training accuracies for superpixel graph classification on MNIST and CIFAR10 \cite{dwivedi2020benchmarking} are higher than the test accuracies.
This phenomenon is especially obvious on the CIFAR10 dataset.
Underfitting refers to the model neither unable to perform well on training data nor generalize well on unseen samples.
Underfitting can be caused by issues such as vanishing and exploding gradients.
Li \emph{et al.} \cite{li2018deeper} identified the oversmoothing issue in training graph networks.
The oversmoothing issue comes with repeatedly applying graph convolutions, resulting in node features across different classes  converging to similar values at deeper layers.
As a result, state-of-the-art graph networks usually adopt a small number of layers \cite{li2019deepgcns}, \emph{e.g.,} 2 to 4.
Further increasing the number of layers will lead to reduced performance.

Regularization has been commonly used  to improve the generalization performance of neural networks.
For graph networks, however, commonly used regularization techniques such as L2 regularization and Dropout \cite{srivastava2014dropout} can only slightly improve the generalization performance \cite{velickovic2018graph}.
Rong \emph{et al.} \cite{rong2020dropedge} proposed the DropEdge method as a variant of Dropout  for regularizing graph  networks.
DropEdge randomly removes a number of edges from the input graph at each training epoch.
It works as a data augmentation method and a message passing reducer.
This method is more effective to improve the performance than Dropout.
Recently, increasing  research attention has been focused on  the oversmoothing issue \cite{zhao2019pairnorm,feng2020graph,xu2018representation,yang2020revisiting}.
Zhao  \emph{et al.} \cite{zhao2019pairnorm} proposed PariNorm, a normalization layer that ensures the total
pairwise feature distance remains to be constant across layers, preventing node features from converging  to similar values.
Feng \emph{et al.} \cite{feng2020graph} proposed  random propagation as a data augmentation method to mitigate oversmoothing.
This method enables the model to perform high-order feature propagation, reducing the risk of node features becoming oversmooth.
While increasing research efforts have been devoted to this issue,
the implication of the oversmoothing issue, \emph{i.e.,} whether it leads to overfitting or underfitting, still remails unknown.
In this paper, we propose a stochastic regularization method to  address the oversmoothing issue.
In our method, we stochastically scale features and gradients (SSFG) in the training procedure.
Our idea is to explicitly apply  stochastic scaling factors to node features to mitigate the feature convergence issue. 
The factors are sampled from a distribution  transformed from the beta distribution.
Applying stochastic scaling to gradients in backward propagation further increases randomness, this is complementary to that applied in forward propagation to improve the overall performance.
We show that by tackling oversmoothing with our SSFG regularization method, both the overfitting issue and the underfitting issue can be reduced.

Our SSFG regularization method can  be seen as a variant  of the Dropout method.
Unlike Dropout, we preserve all neurons and stochastically drop out  or add back a portion of the node feature in forward propagation.
Our method can also be seen as a stochastic rectified linear unit  (ReLU)  function \cite{nair2010rectified} when used together with ReLU.
It generalizes the standard ReLU by using stochastic slops in forward and backward propagations.
We validate our SSFG regularization method on three commonly used graph convolutional networks, \emph{i.e,} Graphsage \cite{hamilton2017inductive}, graph attention networks (GATs) \cite{velickovic2018graph} and gated graph convnets (GatedGCNs) \cite{bresson2017residual}, and conduct experiments on seven benchmark datasets for four graph-based tasks, \emph{i.e.,} graph classification, node classification, link prediction and graph regression.
Extensive experimental results demonstrate that our SSFG regularization is effective in improving the overall performance of the three baseline graph networks.

The contributions of this paper can be summarized as follows:
\begin{itemize}
  \item We propose a stochastic regularization method for graph convolutional networks.
        In our method, we stochastically scale features and gradients in the training procedure.
        As far as we know, this is the first research on regularizing graph networks at both the feature level and the gradient level.
        Our SSFG regularization does not require additional trainable parameters.
        We show that both the overfitting issue and the underfitting issue can be addressed by using the proposed  regularization method.

  \item We experimentally evaluate our SSFG regularization on three types of  commonly used graph networks, \emph{i.e.,} Graphsage, GAT and GatedGCN.
      Extensive experimental results on five benchmark datasets for four graph-based tasks demonstrate that our regularization effectively improves the overall performance of the three baseline graph networks.
\end{itemize}

\section{Related Work}

\subsection{Graph Convolutional Networks}

Graph convolutional networks have become the dominant approach for learning  on graph-structured data.
Existing studies on  graph convolutional networks can be  categorized into two approaches: the spectral-based approach and the spatial-based approach \cite{wu2020comprehensive}.
The spectral-based approach works with the spectral representation of graphs, while the spatial-based approach directly defines convolutions on graph nodes that are spatially close.

Burana \emph{et al.} \cite{bruna2014spectral}  proposed the first spectral-based graph network, in which convolutions are defined in the Fourier domain on the eigen-decomposition of the graph Laplacian.
Defferrard \emph{et al.} \cite{defferrard2016convolutional} later proposed Chebyshev spectral networks (ChebNets) to address the limitation of non-spatially localized filters in Burana's work. 
ChebNets approximate the filters using Chebyshev expansion of the graph Laplacian, resulting in spatially localized filters.
Kips \emph{et al.} \cite{kipf2017semi} further introduced  an efficient layer-wise propagation rule based on the first-order approximation of spectral convolutions.
In spectral-based graph networks, the learned filters depend on the graph structure,
therefore a model trained on a specific graph cannot be applied to other graph structures.

\begin{figure*}[!t]

  \begin{center}
  \hspace{5pt}
  \includegraphics[width=.732\textwidth]{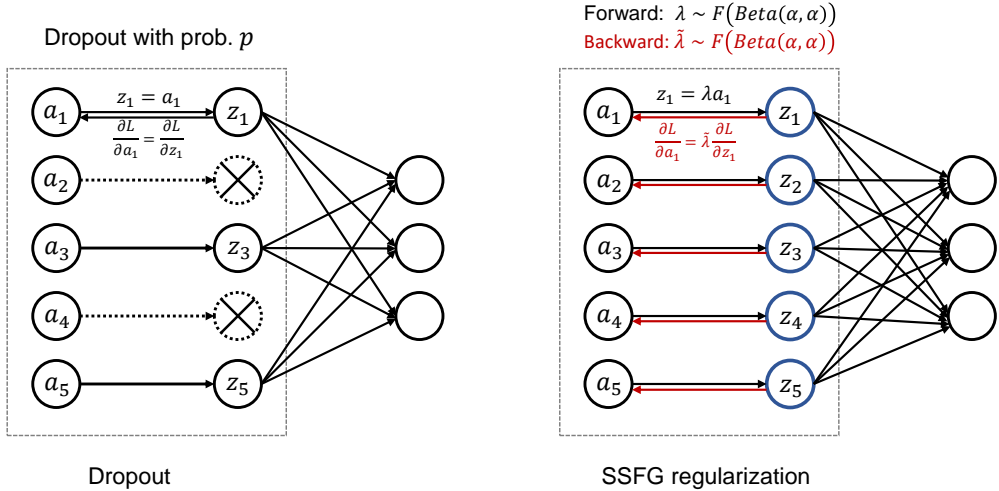} 
  \end{center}
  \caption{An illustration of the SSFG regularization method and its comparison to Dropout. Unlike Dropout, all neurons are preserved  in our SSFG method. The input feature is stochastically scaled in  forward propagation, and the gradient of the output feature is also stochastically scaled in backward propagation. The scaling factors are sampled from a distribution transformed from the Beta function.}
  \label{fig:ssfg}
\end{figure*}

Unlike the spectral-based approach, the spatial-based approach defines convolutions in the spatial domain and  updates node features by aggregating neighbourhood information.
To deal with variable-sized neighbors, several sampling-based methods have been proposed for efficient graph learning.
These methods include the nodewise sampling-based method \cite{hamilton2017inductive}, the layerwise sampling-based method \cite{chen2018fastgcn} and the adaptive layerwise sampling-based method \cite{huang2018adaptive}.
Velickovic \emph{et al.} \cite{velickovic2018graph} proposed graph attention networks, in which the self-attention mechanism is used to compute attention weights in feature aggregation.
Zhang \emph{et al.} \cite{zhang2018gaan} further proposed gated attention networks that apply self-attention to the outputs of multi-attention heads to improve the performance.
Bresson \emph{et al.} \cite{bresson2017residual} proposed residual gated graph convnets, integrating edge gates, residual learning \cite{he2016deep} and batch normalization \cite{ioffe2015batch} into graph networks.

The oversmoothing problem occurs in both spectral-based and spatial-based graph networks.
Li \emph{et al.} \cite{li2018deeper} showed that graph convolution  is  a special form of
Laplacian smoothing and proved that repeatedly applying Laplacian smoothing leads to node features converging to similar values.
Min \emph{et al.} \cite{min2020scattering} proposed to augment conventional GCNs with geometric scattering transforms, which enables band-pass filtering of graph signals to reduce oversmoothing.
More recently, Zhao \emph{et al.} \cite{zhao2019pairnorm} proposed the PairNorm method  to tackle  oversmoothing by ensuring the total pairwise feature distance across layers to be constant.
Chen \emph{et al.} \cite{chen2020measuring} introduced to add a MADGap-based regularizor and  use adaptive edge optimization  to address oversmoothing.

\subsection{Regularization Methods}
Regularization methods have been widely used to improve the generalization performance of neural networks.
Conventional regularization methods include early stopping, \emph{i.e.,} terminating the training procedure when the performance on a validation set stops to improve, Lasso regularization,  weight decay and  soft weight sharing \cite{nowlan1992simplifying}.

Srivastava \emph{et al.}  \cite{srivastava2014dropout} introduced Dropout as a stochastic regularization method to prevent overfitting in neural networks.
The key idea of Dropout is to randomly drop out neurons from the neural network in the training procedure.
Dropout can been seen as  perturbing the feature outputted by a layer by  setting the randomly selected feature points to zero.
The idea behind Dropout has also been adopted in graph networks.
For example in GraphSAGE \cite{hamilton2017inductive}, a fix-sized number of neighbors are sampled for each node in feature aggregation.
This facilitates fast training and is also helpful to improve the overall performance.
The method of node sampling in GraphSAGE can be seen as using random subgraphs of the original graph for training.
Rong \emph{et al.} \cite{rong2020dropedge} proposed the DropEdge method that randomly drops out edges in each training epoch.
This method works as a data augmentor and also a message passing reducer.
It helps to reduce the convergence speed of oversmoothing.
Feng \emph{et al.} \cite{feng2020graph} recently proposed graph random networks which use a random propagation strategy to perform the high-order feature propagation to prevent oversmoothing.

\iftrue
\section{Methodology}
In this section, we first introduce the notations and the oversmoothing issue in graph networks.
Then, we present the proposed SSFG regularization method for tackling the oversmoothing issue, demonstrating its relationship to Dropout and ReLU.
Finally, we introduce the use of our SSFG regularization on three types of commonly used  graph networks.

\begin{figure*}[!t]

  \begin{center}
  \hspace{2pt}\includegraphics[width=.722\textwidth]{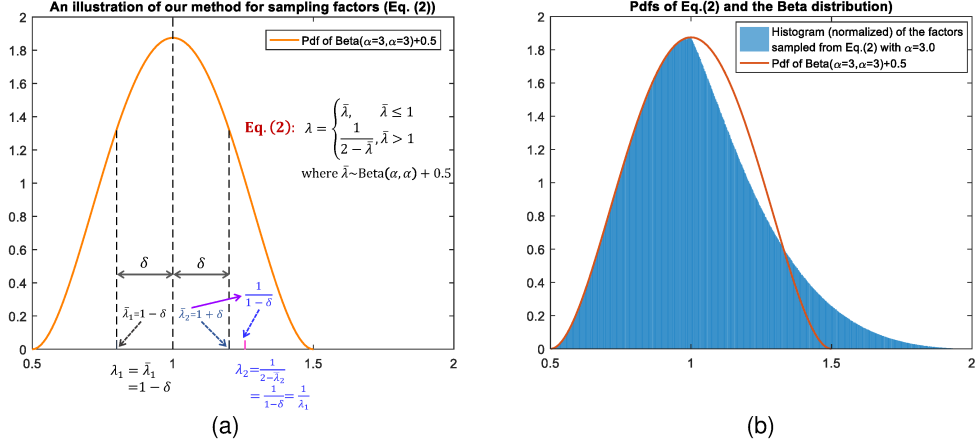} 
  \end{center}
  \caption{(a) An illustration of our method  for sampling scaling factors (see Eq. (\ref{eq:lambda_dist})), note that $p_{Beta+0.5}(\bar{\lambda}_2=1+\delta) = p_{Beta+0.5}(\bar{\lambda}_1=1-\delta)$,  $\lambda_2 = \frac{1}{\lambda_1}$. (b) Probability distribution functions of Eq. (\ref{eq:lambda_dist})) and the shifted Beta distribution, the probability distribution function of Eq. (\ref{eq:lambda_dist})) is shown using the normalized histogram of 10 million sampled factors.}
  \label{fig:pdfeq2}
\end{figure*}

\begin{figure*}[!t]

  \begin{center}
  \includegraphics[width=.702\textwidth]{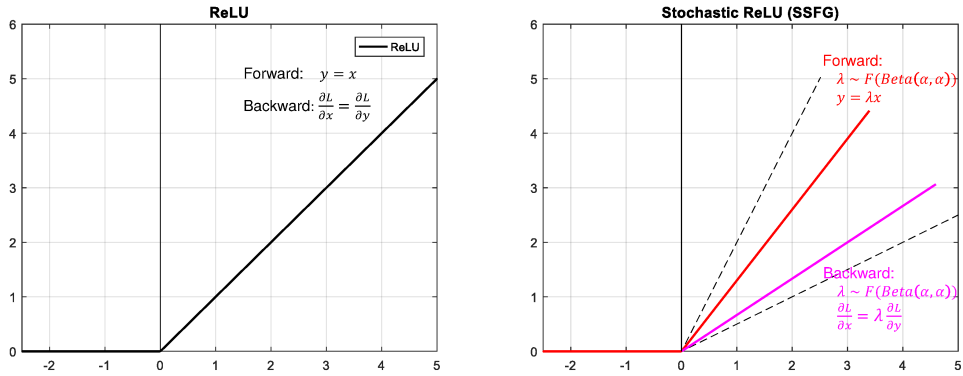} 
  \end{center}
  \caption{Comparison of our SSFG method and ReLU. When used together with ReLU, our SSFG can be seen as a stochastic ReLU, using random slopes in forward and backward propagations.}
  \label{fig:stchrelu}
\end{figure*}

\subsection{Notations}
Let a graph with $N$ nodes be denoted $\mathcal{G} = (V, E, X)$, where $V=\{v_1, v_2,...,v_N\}$ is the node set, $E \subseteq V \times V$ the edge set, and $X \in \mathbb{R}^{N\times C}$ is the feature matrix associated with the nodes.  
$D=\text{diag}(deg_1,...,deg_n) \in \mathbb{R}^{N\times N}$ is the degree matrix, and $A \in \mathbb{R}^{N\times N}$, where $A_{ij}$ equals to $1$ if $v_i$ is connected to $v_j$ or $0$ otherwise, is the adjacent matrix of $\mathcal{G}$.
We consider the nodes of $\mathcal{G}$ to be self-connected, then the structure of $\mathcal{G}$ can be represented as $\tilde{A}=A+I$, where $I$ is the identity matrix.
$\tilde{D}=D+I$ is the augmented degree matrix, and $\tilde{A}_{sym}=\tilde{D}^{-1/2}\tilde{A}\tilde{D}^{-1/2}$ is the symmetrically normalized adjacency matrix of $\tilde{A}$.

\subsection{The Oversmoothing Issue}
While graph networks have achieved state-of-the-art performance for various  graph-based tasks, these models are mostly restricted to shallow layers, \emph{e.g.,} 2 to 4.
Oversmoothing is one of the major issues that restrict the depth of graph networks.
Oversmoothing comes with the nature of graph convolutions which update node features by aggregating neighborhood information.
Repeatedly applying graph convolutions results in node features across different categories converging to similar values regardless of input features, eliminating the discriminative information from these features.

From the view of Zhao \emph{et al.} \cite{zhao2019pairnorm},  oversmoothing  can be understood as follows.
Let ${X}_{\cdot j} \in \mathbb{R}^n$ denote the $j$-th column of $X$, \emph{i.e.,} the input feature of $j$-th node, we can obtain the following:

\begin{equation} \label{eq:lambda_dist}
    \begin{aligned}
      \lim_{k \to +\infty} \tilde{A}_{sym}^{k} \cdot X_{\cdot j} = \pi_j, \hspace{5pt} j=1,2,...,N.
    \end{aligned}
\end{equation}
The normalized representation $\pi=\frac{\pi_j}{\left\| \pi_j \right\|_1}$ satisfies $\pi_i = \frac{\sqrt{deg_i}}{\sum_i\sqrt{deg_i}}$.
Note that $\pi$ is only a function of the graph structure,  regardless of the input feature.
Li \emph{et al.} \cite{li2019deepgcns} point out that oversmmothing leads to the vanishing gradient problem, making graph networks difficult to optimize.


While the oversmoothing issue has been widely discussed, the implication of this issue, \emph{i.e.,} whether it leads to overfitting or underfitting, still remains unsolved.
The studies of \cite{rong2020dropedge,zhao2019pairnorm,yang2020revisiting} regard overfitting and oversmoothing as separate issues.
Yang \emph{et al.} \cite{yang2020revisiting} further suggest that overfitting is a major issue that affects graph networks as compared to oversmoothing.
In this work, we show that by tackling the oversmmothing issue with our SSFG regularization, both the underfitting issue and the overfitting issue can be alleviated.
This indicates that \textbf{oversmmothing can lead to both the overfitting issue and the underfitting issue}.

\subsection{SSFG Regularization}

Dropout randomly drop out neurons, as well as their connections, from the  neural network during training.
Applying Dropout to a neural network can be seen as training many subnetworks of the original  network and  using the ensemble of these subnetworks to make predictions at test time \cite{srivastava2014dropout}.
Although the idea behind Dropout has been applied in graph networks \cite{rong2020dropedge,feng2020graph}, these studies do not directly address the oversmoothing issue.


We introduce SSFG regularization to address the oversmoothing issue.
In SSFG, we stochastically scale features and gradients in the training procedure.
Our idea is to explicitly apply a random scaling factor to break feature convergence.
Specifically,  we multiply each node feature by a factor sampled from a probability distribution in forward propagation.
In backward propagation the gradient of each node feature is also multiplied  by a factor sampled from the probability distribution.
We wish the expectation of the cumulated factors that are applied to node features and gradients to be unchanged.
To this end, we adopt the following trick to define the scaling factor.

\begin{equation} \label{eq:lambda_dist}
    \begin{aligned}
      \lambda &= \begin{cases}
             \bar{\lambda}, \qquad \quad \hspace{7pt} \bar{\lambda} \le 1 \\
                1/(2-\bar{\lambda} ), \ \bar{\lambda} > 1
             \end{cases} \hspace{-10pt}, \\
                 &\text{where} \ \bar{\lambda} \sim \text{Beta}(\alpha, \alpha)+0.5.
    \end{aligned}
\end{equation}
$\text{Beta}(\alpha,\alpha)$ is a probability distribution with the mean equal to 0.5, and $\alpha$ is a hyperparameter.
When $\alpha$ is set to 1.0, $\text{Beta}(\alpha,\alpha)$ is equivalent to the standard uniform distribution.
With the above method, the scaling factor falls in the interval $[0.5,2]$. 
By directly scaling features, the oversmoothing issue is mitigated.
We detail  our SSFG regularization method in Algorithm \ref{alg:spfg}.
It is worth noting that this method does not require additional trainable model parameters.


\begin{algorithm}[tb]
\caption{Pseudocode of our SSFG regularization method in a Pytorch-like style.}
\label{alg:spfg}

        \begin{algorithmic}[1] 
            \Require Node features $\mathbf{h}\in\mathbb{R}^{N\times d}$ ($N$ nodes, each with a $d$ dimensional feature), hyperparameter $\alpha$ used in the Beta function for sampling.
            \Function {Forward}{$\mathbf{h}$}
                \State beta = torch.distributions.beta.Beta($\alpha$, $\alpha$)
                \State lamda = beta.sample($\mathbf{h}$.shape[:1]) + 0.5
                \State lamda[lamda $>$ 1] = 1/(2-lamda[lamda $>$ 1])
                \State return lamda $*$ $\mathbf{h}$
            \EndFunction

            \Function {Backward}{$\mathbf{grad\_output}$}
               \State beta = torch.distributions.beta.Beta($\alpha$, $\alpha$)
                \State lamda = beta.sample($\mathbf{grad\_output}$.shape[:1]) + 0.5
                \State lamda[lamda $>$ 1] = 1/(2-lamda[lamda $>$ 1])
                \State return lamda $*$ $\mathbf{grad\_output}$
%
            \EndFunction

        \end{algorithmic}

\end{algorithm}

A schematic illustration of our SSFG regularization method and its comparison to Dropout is shown in Fig.  \ref{fig:ssfg}.
Unlike the Dropout method, which  performs drop out at the neuron level, our SSFG regularization in forward propagation can be seen as an variant of Dropout that is applied at the feature level.
When the scaling factor $\lambda$ is smaller than $1$, a linear proportion of the node feature is dropped out; and when the scaling factor $\lambda$ is larger than $1$, a linear proportion of the node feature is added back to the node feature.
Applying stochastic scaling to gradients further  increases randomness in the optimization procedure.
This can further helps improve the overall performance.
To the best of our knowledge, this is the first research on regularizing neural networks at both the feature level and the gradient level.
We show through experiments that stochastically scaling gradients is complementary to stochastically scaling features to improve the overall performance.

For node $v_i$, the cumulated factor applied to its hidden feature outputted at layer $l$ can be approximated as follows:
\begin{equation} \label{eq:loss_int}
    \begin{aligned}
    \Lambda = \prod_{i=1}^{l} \lambda_i =\lambda_1\lambda_2\cdots\lambda_l, 
    \end{aligned}
\end{equation}
where $\lambda_1,\lambda_2,...,\lambda_l$ are sampled independently using Eq. (\ref{eq:lambda_dist}).
As shown in Fig. \ref{fig:pdfeq2}, $p(1-\delta)=p(\frac{1}{1-\delta})$ for any $0\le\delta\le0.5$, therefore the expectation of $\Lambda$ equals to $1$, \emph{i.e.,} $\mathbb{E}(\Lambda)=1$.
Based on this analysis, we apply a scaling factor of $1$ to node features at test time for target tasks.
In our experiments, we also compare the performance of using different scaling factors at test time.

ReLU has been commonly used as the nonlinear activation function in graph networks.
When used together with ReLU, our SSFG can be seen as a stochastic ReLU activation function.
An illustration of this explanation is shown in Fig. \ref{fig:stchrelu}.
By using stochastic slopes in both forward and backward propagations, the network model can be robust to different feature variations.
This property makes our SSFG method not specific to graph networks.
It could potentially  be applied to other types neural networks such as (Vision) Transformers \cite{vaswani2017attention,dosovitskiy2020image} and convolutional neural networks to improve the generalization performance.


%
%

\subsection{Regularizing Graph Networks with SSFG}


A typical graph convolutional network takes $X$ and the graph structure $\tilde{A}$ as input and updates node features layerwisely as follows:
\begin{equation} \label{eq:loss_int}
    \begin{aligned}
    h_{i}^{l+1} = f(h_{i}^{l}, \{h_{j}^{l}\}_{j\in \mathcal{N}_i}), \ l=0,...,L-1,
    \end{aligned}
\end{equation}
where $L$ is the number of graph convolutional layers and $\mathcal{N}_i$ is the set of neighbor nodes of  $v_i$.
The proposed SSFG regularization is a general method that can be applied to a wide variety of graph networks.
In this work, we evaluate our SSFG on  three types of commonly used graph convolutional networks, \emph{i.e.,} Graphsage, GAT and GatedGCN, to demonstrate its effectiveness.

\textbf{Graphsage}
Graphsage is a general inductive framework that leverages node feature information  to efficiently generate node features for previously unseen data.
In a Graphsage layer, a  fixed-size set of neighbors are randomly sampled, and the features of these sampled neighbors are aggregated using an  aggregator function, such as the mean operator and LSTM \cite{hochreiter1997long} to update a node's feature as follows:
\begin{equation} \label{eq:loss_int}
    \begin{aligned}
    h_{i}^{l+1} = \sigma(W  \cdot \text{Concat}( h_{i}^{l}, \text{Aggregator}\{h_{j}^{l}\}_{j\in \mathcal{N}_i})),
    \end{aligned}
\end{equation}
where $W$ is the weight matrix of the shared linear transformation function, and $\sigma$ is a nonlinear activation function, \emph{e.g.,} ReLU activation function.
SSFG can be applied to input node features to the Graphsage  layer or after the nonlinear activation function.

\textbf{GATs}
GATs are inspired by the self-attention mechanism \cite{vaswani2017attention} that are widely applied in NLP and computer vision tasks.
In a GAT layer, each neighbor of a  node is assigned with an attention weight computed by an attention function in feature aggregation as follows:
\begin{equation} \label{eq:loss_int}
    \begin{aligned}
    h_{i}^{l+1} = \sigma \left( \sum_{j \in \mathcal{N}_i} \text{attn}(h_{i}^{l},h_{j}^{l}) \cdot W  \cdot h_{j}^{l} \right),
    \end{aligned}
\end{equation}
where {attn} is the attention function.
GATs usually employs multiple attention heads to improve the overall performance.
By default, we apply our SSFG regularization to the output of each attention head.

\textbf{GatedGCNs}
GatedGCNs use the edge gating mechanism \cite{marcheggiani2017encoding} and residual connection in aggregating features from a node's  local neighborhood as follows:
\begin{equation} \label{eq:loss_int}
    \begin{aligned}
    h_{i}^{l+1} = \sigma \left(U  \cdot h_i^l + \sum_{j \in \mathcal{N}_i} e_{ij}^l \odot V  \cdot h_j^l \right) + h_j^l, \\
    e_{ij}^{l+1} = \sigma(A \cdot h_i^{l} + B  \cdot h_j^{l}) + e_{ij}^{l}, \hspace{52pt}
    \end{aligned}
\end{equation}
where $U,V,A,B$ are weight matrices of linear transformations, and $\odot$ is the Hadamard product.
GatedGCNs explicitly maintain edge feature $e_{ij}$ at each layer.
By default, we apply our SSFG regularization to both the node features and the edge features outputted by a GatedGCN layer.

\section{Experiments}

\begin{table}[tbp]
\caption{Statistics of the seven benchmark datasets used in our experiments. On COLLAB, edges that represent collaborations up to 2017 are used for training, and edges that represent collaborations in 2018 and 2019 are used for validation and testing, respectively.   }

\centering  
\begin{tabular}{ l|cccccc }
\toprule[1pt]
Dataset  &Graphs & Nodes/graph &$\#$Training &$\#$Val. &$\#$Test \\
\midrule[.8pt]

PATTERN & 14K &44-188 &10,000 &2000 &2000\\
CLASTER & 12K &41-190 &10,000 &1000 &1000\\

\hline

MNIST   &70K &40-75  &55,000 &5000 &10,000 \\
CIFAR10 &60K &85-150 &45,000 &5000 &10,000\\
\hline
TSP    &12K &50-500  &10,000 &1000 &1000\\
COLLAB &1   &235,868 & - & - & -\\
\hline
ZINC &12K &9-37 &10,000 &1000 &1000\\
\hline

\end{tabular}

\label{table:statics}
\end{table}

\begin{table*}[tbp]
\caption{Node classification results on PATTERN and CLUSTER. We experiment using 4 and 16 layers in the graph networks.}

\begin{minipage}[b]{.48\linewidth}
\centering  
\begin{tabular}{l|c|ccccc }
\toprule[1pt]

\multirow{2}{*}{ Method}   &\multirow{2}{*}{$L$} & \multicolumn{2}{c}{PATTERN}  \\
& & Test (Acc.) & Train (Acc.)  \\
\midrule[.6pt]
Graphsage w/o SSFG &\multirow{2}{*}{4} & 50.516$\pm$0.001 &50.473$\pm$0.014  \\
Graphsage + SSFG ($\alpha$=+$\infty$) & & 50.516$\pm$0.001 &50.473$\pm$0.014\\
\hline
Graphsage w/o SSFG &\multirow{2}{*}{16} &50.492$\pm$0.001 &50.487$\pm$0.005 \\
Graphsage + SSFG ($\alpha$=+$\infty$) & &50.492$\pm$0.001 &50.487$\pm$0.005 \\
\midrule[.6pt]

GAT w/o SSFG &\multirow{2}{*}{4} &75.824$\pm$1.823 &77.883$\pm$1.632 \\
GAT + SSFG ($\alpha$=8.0) & &\textbf{77.290$\pm$0.469} &\textbf{77.938$\pm$0.528} \\
\hline
GAT w/o SSFG &\multirow{2}{*}{16} &78.271$\pm$0.186 &90.212$\pm$0.476\\
GAT + SSFG ($\alpha$=8.0) & &\textbf{81.461$\pm$0.123} &\textbf{82.724$\pm$0.385}\\
\midrule[.6pt]

GatedGCN w/o SSFG &\multirow{5}{*}{4} &84.480$\pm$0.122 &84.474$\pm$0.155 \\
GatedGCN + SSFG ($\alpha$=3.0) & &85.205$\pm$0.264 &85.283$\pm$0.347 \\
GatedGCN + SSFG ($\alpha$=4.0) & &85.016$\pm$0.181 &84.923$\pm$0.202 \\
GatedGCN + SSFG ($\alpha$=5.0) & &\textbf{85.334$\pm$0.175} &\textbf{85.316$\pm$0.192} \\
GatedGCN + SSFG ($\alpha$=6.0) & &85.102$\pm$0.161 &85.066$\pm$0.155 \\

\hline

GatedGCN w/o SSFG &\multirow{3}{*}{16} &85.568$\pm$0.088 &86.007$\pm$0.123 \\
GatedGCN + SSFG ($\alpha$=3.0) & &\textbf{85.723$\pm$0.069} &\textbf{85.625$\pm$0.072} \\
GatedGCN + SSFG ($\alpha$=4.0) & &85.717$\pm$0.020 &85.606$\pm$0.012 \\
GatedGCN + SSFG ($\alpha$=5.0) & &85.651$\pm$0.054 &85.595$\pm$0.048 \\
\bottomrule[.6pt]

\end{tabular}
\end{minipage}
\begin{minipage}[b]{.48\linewidth}
\centering  
\begin{tabular}{l|c|ccccc }
\toprule[1pt]

\multirow{2}{*}{ Method}   &\multirow{2}{*}{$L$} & \multicolumn{2}{c}{CLUSTER}  \\
& & Test (Acc.) & Train (Acc.)  \\

\midrule[.6pt]

Graphsage w/o SSFG &\multirow{2}{*}{4} &50.454$\pm$0.145 &54.374$\pm$0.203  \\
Graphsage + SSFG ($\alpha$=5.0) & &\textbf{50.562$\pm$0.070} &\textbf{53.014$\pm$0.025}\\
\hline
Graphsage vanilla &\multirow{2}{*}{16} &63.844$\pm$0.110 &86.710$\pm$0.167 \\
Graphsage + SSFG ($\alpha$=7.0) & &\textbf{66.851$\pm$0.066} &\textbf{79.220$\pm$0.023}\\
\midrule[.6pt]

GAT w/o SSFG &\multirow{2}{*}{4} &57.732$\pm$0.323 &58.331$\pm$0.342 \\
GAT + SSFG ($\alpha$=7.0) & &\textbf{59.888$\pm$0.044} &\textbf{59.656$\pm$0.025}\\
\hline
GAT w/o SSFG &\multirow{2}{*}{16} &70.587$\pm$0.447 &76.074$\pm$1.362 \\
GAT + SSFG ($\alpha$=4.0) & &\textbf{73.689$\pm$0.088} &\textbf{79.476$\pm$0.302}\\
\midrule[.6pt]

GatedGCN w/o SSFG &\multirow{5}{*}{4} & 60.404$\pm$0.419 &61.618$\pm$0.536\\
GatedGCN + SSFG ($\alpha$=6.0) & &61.028$\pm$0.302 &62.415$\pm$0.311 \\
GatedGCN + SSFG ($\alpha$=7.0) & &61.222$\pm$0.267 &62.844$\pm$0.352 \\
GatedGCN + SSFG ($\alpha$=8.0) & &\textbf{61.498$\pm$0.267} &\textbf{63.310$\pm$0.343} \\
GatedGCN + SSFG ($\alpha$=9.0) & &61.375$\pm$0.047 &63.049$\pm$0.134 \\

\hline

GatedGCN w/o SSFG &\multirow{3}{*}{16} &73.840$\pm$0.326 &87.880$\pm$0.908 \\
GatedGCN + SSFG ($\alpha$=4.0) & &75.671$\pm$0.084 &83.769$\pm$0.035 \\
GatedGCN + SSFG ($\alpha$=5.0) & &\textbf{75.960$\pm$0.020} &\textbf{83.623$\pm$0.652} \\
GatedGCN + SSFG ($\alpha$=6.0) & &75.601$\pm$0.078 &84.516$\pm$0.299 \\
\bottomrule[.6pt]

\end{tabular}
\end{minipage}

\label{table:nodecls}
\end{table*}

\begin{table*}[tbp]
\caption{Superpixel graph classification results on MINIST and CIFAR10.  The number of  layers is set to 4.}
\begin{minipage}[b]{.48\linewidth}
\centering  
\begin{tabular}{l|ccccc }
\toprule[1pt]

\multirow{2}{*}{ Method}   & \multicolumn{2}{c}{MNIST}  \\
 & Test (Acc.) & Train (Acc.)  \\
\midrule[.6pt]
Graphsage vanilla  & 97.312$\pm$0.097 &100.00$\pm$0.000  \\
Graphsage + SSFG ($\alpha$=5.0) &\textbf{97.943$\pm$0.147} &\textbf{99.996$\pm$0.002}  \\
\midrule[.6pt]

GAT vanilla & 95.535$\pm$0.205 &99.994$\pm$0.008  \\
GAT + SSFG ($\alpha$=2.0) &\textbf{97.938$\pm$0.075} &\textbf{99.996$\pm$0.002}  \\
\midrule[.6pt]

GatedGCN vanilla & 97.340$\pm$0.143 &100.00$\pm$0.000\\
GatedGCN + SSFG ($\alpha$=1.0) &97.848$\pm$0.106 &99.889$\pm$0.035\\

GatedGCN + SSFG ($\alpha$=1.5) &97.730$\pm$0.116 &99.975$\pm$0.004 \\
GatedGCN + SSFG ($\alpha$=2.0) &\textbf{97.985$\pm$0.032} &\textbf{99.996$\pm$0.001}  \\
GatedGCN + SSFG ($\alpha$=2.5) &97.703$\pm$0.054 &99.996$\pm$0.001  \\

\bottomrule[.6pt]

\end{tabular}
\end{minipage}
\begin{minipage}[b]{.48\linewidth}
\centering  
\begin{tabular}{l|ccccc }
\toprule[1pt]

\multirow{2}{*}{ Method}   & \multicolumn{2}{c}{CIFAR10}  \\
 & Test (Acc.) & Train (Acc.)  \\
\midrule[.6pt]
Graphsage vanilla  &65.767$\pm$0.308 &99.719$\pm$0.062  \\
Graphsage + SSFG ($\alpha$=4.0) &\textbf{68.803$\pm$0.471} &\textbf{89.845$\pm$0.166} \\
\midrule[.6pt]

GAT vanilla & 64.223$\pm$0.455 &89.114$\pm$0.499\\
GAT + SSFG ($\alpha$=4.0) &\textbf{66.065$\pm$0.171}  &\textbf{84.383$\pm$0.986} \\
\midrule[.6pt]

GatedGCN vanilla & 67.312$\pm$0.311  &94.553$\pm$1.018\\
GatedGCN + SSFG ($\alpha$=1.0) &71.585$\pm$0.361 &83.878$\pm$1.146 \\

GatedGCN + SSFG ($\alpha$=1.5) &\textbf{71.938$\pm$0.190} &\textbf{87.473$\pm$0.593}  \\
GatedGCN + SSFG ($\alpha$=2.0) &71.383$\pm$0.427 &87.745$\pm$0.973 \\
GatedGCN + SSFG ($\alpha$=2.5) &70.913$\pm$0.306 &88.645$\pm$0.750\\

\bottomrule[.6pt]
\end{tabular}
\end{minipage}

\label{table:graphcls}
\end{table*}

\begin{table*}[tbp]
\caption{Link prediction results on TSP and COLLAB. The number of  layers is set to 4. }
\begin{minipage}[b]{.48\linewidth}
\centering  
\begin{tabular}{l|ccccc }
\toprule[1pt]

\multirow{2}{*}{ Method}   & \multicolumn{2}{c}{TSP}  \\
 & Test (F1) & Train (F1)  \\
\midrule[.6pt]
Graphsage w/o SSFG  & 0.665$\pm$0.003 &0.669$\pm$0.003  \\
Graphsage + SSFG ($\alpha$=5.0) &\textbf{ 0.714$\pm$0.003} &\textbf{0.717$\pm$0.003} \\
\midrule[.6pt]

GAT w/o SSFG &0.671$\pm$0.002 &0.673$\pm$0.002 \\
GAT + SSFG ($\alpha$=300) &\textbf{0.682$\pm$0.000} &\hspace{.8pt} \textbf{0.684$\pm$0.0001} \\
\midrule[.6pt]

GatedGCN w/o SSFG  &0.791$\pm$0.003 &0.793$\pm$0.003 \\
GatedGCN + SSFG ($\alpha$=4.0) &0.802$\pm$0.001 &0.804$\pm$0.001 \\

GatedGCN + SSFG ($\alpha$=5.0) &\textbf{0.806$\pm$0.001} &\textbf{0.807$\pm$0.001} \\
GatedGCN + SSFG ($\alpha$=6.0) &0.805$\pm$0.001 &0.808$\pm$0.001 \\
GatedGCN + SSFG ($\alpha$=7.0) &0.805$\pm$0.001 &0.807$\pm$0.001  \\
\bottomrule[1.pt]

\end{tabular}
\end{minipage}
\begin{minipage}[b]{.48\linewidth}

\begin{tabular}{l|ccccc }

\midrule[1.pt]

\multirow{2}{*}{ Method}   & \multicolumn{2}{c}{COLLAB}  \\
 & Test (Hits@50) & Train (Hits@50)  \\
\midrule[.6pt]
Graphsage w/o SSFG  &51.618$\pm$0.690 &99.949$\pm$0.052  \\
Graphsage + SSFG ($\alpha$=4.0) &\textbf{53.146$\pm$0.230} &\textbf{98.280$\pm$1.300} \\
\midrule[.6pt]

GAT w/o SSFG &51.501$\pm$0.962 &97.851$\pm$1.114\\
GAT + SSFG ($\alpha$=4.0) &53.616$\pm$0.400 &97.700$\pm$0.132\\
GAT + SSFG ($\alpha$=5.0) &53.908$\pm$0.253 &97.835$\pm$0.182 \\
GAT + SSFG ($\alpha$=6.0) &\textbf{54.715$\pm$0.069} &\textbf{97.929$\pm$0.189}  \\
GAT + SSFG ($\alpha$=7.0) &54.252$\pm$0.092 &98.084$\pm$0.340\\
\midrule[.6pt]

GatedGCN w/o SSFG &52.635$\pm$1.168 &96.103$\pm$1.876 \\

GatedGCN + SSFG ($\alpha$=5.0) &\textbf{53.055$\pm$0.671} &\textbf{92.535$\pm$0.989} \\
\bottomrule[1pt]
\end{tabular}
\end{minipage}

\label{table:linkpred}
\end{table*}

\subsection{Experimental Setup}
\subsubsection{Datasets}
Our experiments are conducted on seven recently released  benchmark datasets \cite{dwivedi2020benchmarking}, \emph{i.e.,} PATTERN, CLUSTER, MNIST, CIFAR10, TSP, COLLAB and ZINC.
These datasets are used for four graph-based tasks: node classification (PATTERN, CLUSTER), graph classification (MNIST, CIFAR10), link prediction (TSP, COLLAB) and graph regression (ZINC).
The statistics of the seven datasets are given in Table \ref{table:statics}.

\subsubsection{Training Details}
We closely follow the experimental setup used in the work of Dwivedi \emph{et al.} \cite{dwivedi2020benchmarking}.
The Adam method \cite{kingma2014adam} is used to train all the models.
The learning rate is initialized to $10^{-3}$ and reduced by a factor of 2 if the loss has not improved for a number of epochs (10, 20 or 30), note that we use a larger patience number for some experiments.
The training procedure is terminated when the learning rate is reduced to smaller than $10^{-6}$.
For the node classification task, we  conduct experiments using different layers (4 and 16).
For the remaining tasks, the number of layers is set to 4.
For each evaluation, we run the experiment 4 times using different random seeds and report the mean and standard deviation of the 4 results.
Our method is implemented using  Pytorch \cite{paszke2017automatic} and the DGL library \cite{wang2019deep}.

\subsubsection{Evaluation Metrics}
Following \cite{dwivedi2020benchmarking}, the following evaluation metrics are used for different tasks.
\begin{itemize}
  \item \textbf{Accuracy}. Weighted average node classification accuracy is used for the node classification task (PATTERN and CLUSTER), and classification accuracy is used  for the graph classification task (MNIST and CIFAR10).
  \item  \textbf{F1 score} for the positive class is used for performance  evaluation on the TSP dataset, due to high class imbalance, \emph{i.e.,} only the edges in the TSP tour are labeled as positive.
  \item \textbf{Hits@K} \cite{hu2020open} is used for the  COLLAB dataset, aiming to measure the model's ability to predict future collaboration relationships.
      This method ranks each true collaboration against 100,000 randomly sampled negative collaborations and counts the ratio of positive edges that are ranked at $K$-th place or above.
  \item \textbf{MAE} (mean absolute error) is used to evaluate graph regression performance on ZINC.
\end{itemize}

\begin{figure*}[!t]

  \begin{center}
  \includegraphics[width=\textwidth]{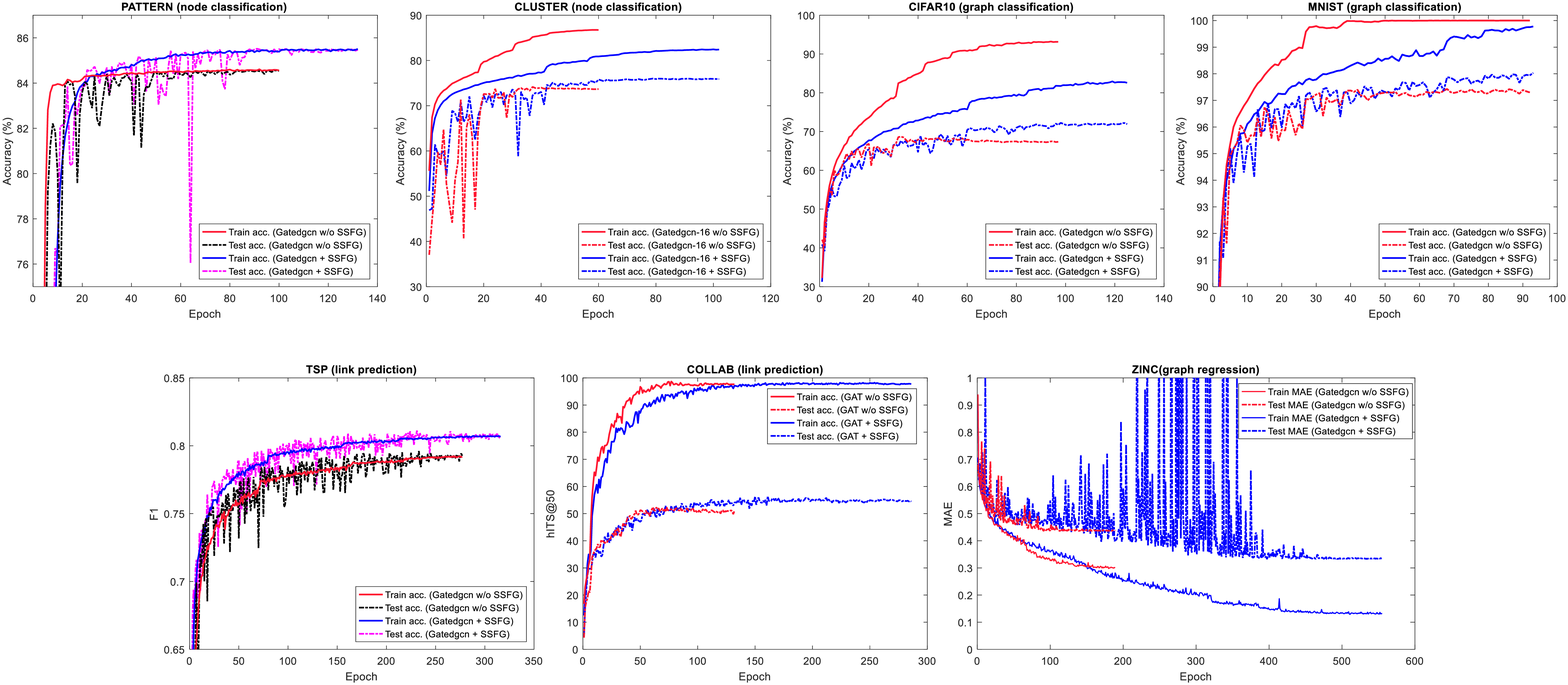} 
  \end{center}
  \caption{Training and test accuracy/F1/Hits@50/MAE curves with respective to the training epoch, showing that our SSFG regularization helps to address both the overfitting issue and the underfitting issue. }
  \label{fig-curve}
\end{figure*}

\subsection{Experimental Results}
\subsubsection{Quantitative Results}

Table \ref{table:nodecls} reports the quantitative results of node classification on PATTERN and CLUSTER.
It can be seen that applying our SSFG regularization effectively improves the test accuracies except for  Graphsage on the PATTERN dataset.
Applying our SSFG regularization on GATs with 16 layers yields 3.190\% and 3.102\% performance improvements on PATTERN and CLUSTER, respectively.
These improvements are higher than those obtained for GATs with 4 layers.
For GatedGCNs, applying our SSFG regularization yields more performance improvements on CLUSTER than those on PATTERN.




The graph classification results on MNIST and CIFAR10 are reported in Table \ref{table:graphcls}.
We see that applying our SSFG regularization  helps improve the overall performance for the three baseline graph networks on the two datasets.
While vanilla Graphsage and GatedGCN perform well compared to GAT on MNIST, the three baseline graph networks with our SSFG regularization achieve comparable performances.
On CIFAR10, applying our SSFG regularization to GatedGCN improves the accuracy from 67.312\% to 71.938, yielding a 4.626\% performance gain, which is higher than those obtained by applying SSFG to Graphsage and GAT.


The results for link prediction  are reported in Table \ref{table:linkpred}.
Once again, applying SSFG regularization results in improved performance for the three baseline graph networks.
On TSP, the use of SSFG regularization in Graphsage, GAT and GatedGCN yields 0.049, 0.011 and 0.015 performance gains, respectively.
On the COLLAB dataset, applying SSFG regularization to Graphsage and GAT yields 2.528 and 3.214 performance improvements, respectively.
For GatedGCN, SSFG slightly improves the prediction performance.

%
%
%
%

Table \ref{table:zinc} reports the experimental results on ZINC, demonstrating the effectiveness of our SSFG regularization to improve  the graph regression performance for the three baseline graph networks.
For Graphsage, GAT and GatedGCN, the use of our SSFG regularization reduces 0.027, 0.009 and 0.098 mean absolute errors, respectively.

We have shown that the proposed SSFG regularization method helps improve the overall performance of the three baseline graph networks for different graph-based tasks.
For most experiments, the performance on the test data improves while that on the training data reduces.
This indicates that the improvements are obtained through reducing the overfitting issue.
It is worth noting that for some experiments, \emph{e.g.,} the experiments on TSP and the experiments of GatedGCN with 4 layers on PATTERN and CLUSTER, the performances on the test data and the training data improve simultaneously.
This indicates that our SSFG regularization also helps to address the underfitting issue.
While the implication of oversmoothing remains unsolved in previous studies, \textbf{our results demonstrate that oversmoothing can lead to both the overfitting issue and the underfitting issue}.

\begin{table}[tbp]
\caption{Graph regression results on ZINC. The number of layers is set to 4.}

\centering  

\small
\begin{tabular}{ >{\footnotesize}l|ccccc }
\toprule[1pt]

\multirow{2}{*}{ Method}   & \multicolumn{2}{c}{ZINC}  \\
 & Test (MAE $\downarrow$) & Train (MAE $\downarrow$)  \\
\midrule[.6pt]
Graphsage vanilla              & 0.468$\pm$0.003 &0.251$\pm$0.004  \\
Graphsage + SSFG ($\alpha$=10) & \textbf{0.441$\pm$0.006} &\textbf{0.191$\pm$0.005 } \\

\midrule[.6pt]
GAT vanilla              & 0.475$\pm$0.007 &0.317$\pm$0.006\\
GAT + SSFG ($\alpha$=20) & \textbf{0.466$\pm$0.001} &\textbf{0.329$\pm$0.010}\\
\midrule[.6pt]

GatedGCN vanilla    & 0.435$\pm$0.011 &0.287$\pm$0.014\\
GatedGCN + SSFG ($\alpha$=100) &0.371$\pm$0.018 &0.175$\pm$0.015 \\
GatedGCN + SSFG ($\alpha$=200) &0.352$\pm$0.023 &0.164$\pm$0.022 \\
GatedGCN + SSFG ($\alpha$=250) &\textbf{0.347$\pm$0.023} &\textbf{0.154$\pm$0.021} \\
GatedGCN + SSFG ($\alpha$=300) &0.357$\pm$0.021 &0.157$\pm$0.010 \\

\bottomrule[.6pt]

\end{tabular}

\label{table:zinc}
\end{table}

\begin{table*}[tbp]
\caption{Ablation study. We show the effect of stochastically scaling features and gradients on the overall performance. }

\centering  

\begin{tabular}{l|cc cc cc|c }
\toprule[1pt]
\multirow{3}{*}{Method}  & \multicolumn{6}{c|}{GatedGCN} & GAT \\
   \cline{2-8}
  & PATTERN &CLUSTER  &MNIST &CIFAR10 &TSP (F1) &ZINK (MAE $\downarrow$)  &COLLAB \\
  &($L$=4, $\alpha$=5.0) & ($L$=16, $\alpha$=5.0) & ($L$=4, $\alpha$=2.0) & ($L$=4, $\alpha$=1.5) & ($L$=4, $\alpha$=5.0)  &($L$=4, $\alpha$=7.0) & {\footnotesize($L$=4, $\alpha$=250)} \\
\midrule[.8pt]

w/o SSFG &84.480 &73.840 &97.340 & 67.312 &0.791  &0.435 &51.501\\
\cline{1-8}
                    +SSF  (forward regu. only) &85.122 &75.518 &97.753 &71.107 &0.802  &0.352 &54.394\\
                    +SSG (backward regu. only) &84.639 &72.951 &97.537 &68.325 &0.795  &0.385 &52.440\\
                    +SSFG (full) &85.334 & 75.960 &97.985 &71.983 & 0.806  &0.347 &54.715\\
\bottomrule[1pt]

\end{tabular}

\label{table:ablation}
\end{table*}

\begin{table*}[tbp]
\caption{Performance comparison of Dropout and our SSFG regularization method. Dropout ($p=0.5$) results in only slightly improved or reduced performance, whereas our SSFG method consistently improves the overall performance.}

\centering  
%
%
%
%
%

\begin{tabular}{l|cc cc cc|c }
\toprule[1pt]
\multirow{3}{*}{Method}  & \multicolumn{6}{c|}{GatedGCN} & GAT \\
   \cline{2-8}
  & PATTERN &CLUSTER  &MNIST &CIFAR10 &TSP (F1) &ZINK (MAE $\downarrow$)  &COLLAB \\
  &($L$=4, $\alpha$=5.0) & ($L$=16, $\alpha$=5.0) & ($L$=4, $\alpha$=2.0) & ($L$=4, $\alpha$=1.5) & ($L$=4, $\alpha$=5.0)  &($L$=4, $\alpha$=7.0) & ($L$=4, $\alpha$=250) \\
\midrule[.8pt]

w/o SSFG/Dropout &84.480 &73.840 &97.340 & 67.312 &0.791  &0.435 &51.501\\
\cline{1-8}
                    +Dropout ($p$=0.5) &50.000 &24.988 &97.494 &67.345 &0.551  &0.675 &51.589\\

                    +SSFG (full) &85.334 & 75.960 &97.985 &71.983 & 0.806  &0.347 &54.715\\
\bottomrule[1pt]

\end{tabular}
\label{table:compwDropout}
\end{table*}

Besides, we observe that for most experiments our SSFG regularization method results in small standard deviations.
On CLUSTER and COLLAB, the use of our SSFG regularization consistently results in small standard deviations  compared to those obtained without using SSFG regularization.
For the remaining datasets, our SSFG regularization also results in small standard  deviations for most cases.
This shows that the proposed SSFG regularization method can stabilize the  learning algorithms.

For the graph networks that achieve the highest performance improvements in different tasks, we also show the impact of the value of $\alpha$ used for sampling scaling factors on the overall performance in the quantitative results.
We see that the value of $\alpha$ has different impacts for different tasks.
Even for the same task, a graph network with different layers may use different values of $\alpha$ to obtain the best task performance.
This indicates that the value of $\alpha$ needs to be carefully tuned for the best task performance.

Fig. \ref{fig-curve} shows the training and test accuracy curves with respect to the training epoch.
We see that our SSFG regularization method helps to address both the overfitting issue and the underfitting issue.
Compared with the work of Dwivedi \cite{dwivedi2020benchmarking}, we use a large patience value for the optimizer when learning on some datasets.
Therefore, it can take more epochs to complete the training procedure.
We also observe that the training procedure takes comparable epochs on some datasets (\emph{e.g.,} MNIST), as that without using SSFG regularization.
As aforementioned, our method can be seen as a stochastic ReLU activation function, the results show that our method outperforms standard ReLU in graph representation learning.
As with the ReLU activation function, our method does not increase the number of learnable parameters.

\subsubsection{Ablation Study}
Our SSFG regularization method performs stochastic scaling at both the feature level and the gradient level.
We conduct an ablation study to show the effect of scaling features and gradients on the overall performance.
In our experiments, we use GAT on COLLAB and GatedGCN on the remaining datasets.
According to the quantitative results, the value of $\alpha$ that achieves the best task performance is used in the experiments.
The ablation study results are reported in Table \ref{table:ablation}.
We see that scaling features and scaling gradients contribute differently on different datasets.
On a whole, stochastically scaling features and gradients are complementary to each other to achieve the best task performance.

\begin{table}[tbp]
\caption{Performance comparison of using different scaling factors at test time for superpixel graph classification  on MNIST and CIFAR10. }

\centering  
\begin{tabular}{ c|ccc } 
\toprule[1pt]
\multirow{2}{*}{Scaling factor}   &\multicolumn{2}{c}{GatedGCN}  \\
 & MNIST ($\alpha$=2.0) &CIFAR10 ($\alpha$=1.5)\\
\midrule[.8pt]
0.8  &97.626$\pm$0.064 &71.377$\pm$0.304 \\
0.9 &97.810$\pm$0.039 &71.637$\pm$0.267 \\

\textbf{1.0}  &\textbf{97.985$\pm$0.032} &\textbf{71.938$\pm$0.190} \\
1.1  &97.759$\pm$0.069 &71.690$\pm$0.244\\
1.2 &97.711$\pm$0.020 &71.441$\pm$0.369 \\
\bottomrule[1pt]

\end{tabular}

\label{table:scalingfactor}
\end{table}

\textbf{Comparison with Dropout}
Our SSFG method can be seen as a variant of Dropout that is applied at both the feature level and the gradient level.
To demonstrate the advantage of our method for graph networks, we further compare the performance of our method and Dropout.
We apply Dropout with probability 0.5 to node features.
The comparison of results are shown in Table \ref{table:compwDropout}.
Dropout results in  slightly improved performance on MINST, CIFAR10 and COLLAB, and reduced performance on the remaining datasets.
In contrast to Dropout, our SSFG regularization consistently improves the overall performance on the seven datasets.
This demonstrates the advantage of our method over Dropout for graph networks.
We also tested using different probability values (0.8 and 0.9) in Dropout, we found that this leads to similar or reduced  performance.
We also found that Dropout requires more epochs to converge than our SSFG method.
Dropout essentially uses the ensemble of numerous trained subnetworks to make a prediction, which does not effectively address the oversmoothing problem; whereas our SSFG regularization addresses the oversmoothing issue to improve the overall performance.

\textbf{Impact of the Scaling Factor at Test Time}
At test time, the scaling factor of 1 is applied to node features.
To show the effectiveness of this approach, we further compare the performance of using difference factors.
Our experiments are conducted on MNIST and CIFAR10, the results are shown in Table \ref{table:scalingfactor}.
We find that using other factors (0.8, 0.9, 1.1 and 1.2) results in reduced performance compared to that obtained using 1.0.
This shows that not applying scaling at time test is a good choice.

\subsection{Broader Impact}
We have shown that our SSFG regularization method is effective in improving the overall performance for  graph networks.
The proposed SSFG regularization method helps to address both the overfitting issue and the underfitting issue without increasing the number of trainable parameters.
When used together with ReLU, our SSFG method can be seen as a stochastic ReLU activation function that is applied at both the feature level and the gradient level.
This explanation makes our SSFG method not specific for graph networks.
Overfitting and underfitting are also issues with neural networks for other tasks such as image recognition and natural language processing tasks.
It could be potentially useful replacing the  standard ReLU  with our SSFG method in the network models to improve the generalization performance, especially when training data are small.



\section{Conclusions}
In this paper, we presented a stochastic regularization method for graph convolutional networks.
In our method, we stochastically scale features and gradients  by a factor sampled from a probability distribution in the training procedure.
Our SSFG regularization method  helps to address the oversmoothing issue caused by repeatedly applying graph convolutional layers.
We showed that applying stochastic scaling at the feature level is complementary to that applied at the gradient level in improving the overall performance.
When used together with ReLU,  our method can also be seen as a stochastic ReLU activation function.
We experimentally validated our SSFG regularization method on seven benchmark datasets for different graph-based tasks, including node classification, graph classification, link prediction and graph regression.
We conducted an ablation study to  show the effects of applying scaling at the feature level and the gradient level on the overall performance.
The experimental results  demonstrated that our SSFG method  helps to address both the overfitting issue and the underfitting issue.
While the oversmoothing issue was identified several years ago, the implication of this issue remains  unsolved.
Our experimental results suggest that the oversmoothing issue can lead to both overfiting and underfitting.

\section{Acknowledgement}
We would like to thank the reviewers for reviewing our manuscript.

\fi

\bibliographystyle{IEEEtran}
\bibliography{mybib}

\end{document}